\documentclass{article}
\usepackage{spconf,amsmath,graphicx}
\usepackage{enumitem}
\usepackage{bm}
\usepackage{amsmath}
\usepackage{amssymb}

\DeclareMathOperator*{\argmin}{arg\,min}

\title{A Coarse-to-fine Framework for Learned Color Enhancement with Non-local Attention}
\name{Chaowei Shan \qquad Zhizheng Zhang \qquad Zhibo Chen* \thanks{*Corresponding author: Zhibo Chen (chenzhibo@ustc.edu.cn). This work was supported in part by NSFC under Grant 61571413, 61632001.}}
\address{CAS Key Laboratory of Technology in Geo-spatial Information Processing and Application System \\ University of Science and Technology of China, Hefei 230027, China}
\begin{document}
%
\maketitle
%
\begin{abstract}

Automatic color enhancement is aimed to adaptively adjust photos to expected styles and tones. For current learned methods in this field, global harmonious perception and local details are hard to be well-considered in a single model simultaneously. To address this problem, we propose a coarse-to-fine framework with non-local attention for color enhancement in this paper. Within our framework, we propose to divide enhancement process into channel-wise enhancement and pixel-wise refinement performed by two cascaded Convolutional Neural Networks (CNNs). In channel-wise enhancement, our model predicts a global linear mapping for RGB channels of input images to perform global style adjustment. In pixel-wise refinement, we learn a refining mapping using residual learning for local adjustment. Further, we adopt a non-local attention block to capture the long-range dependencies from global information for subsequent fine-grained local refinement. We evaluate our proposed framework on the commonly using benchmark and conduct sufficient experiments to demonstrate each technical component within it.

\end{abstract}
\section{Introduction}
\label{sec:intro}
\vspace{-0.5em}
Photo retouching using professional software like Adobe Photoshop or Lightroom is inconvenient for amateur photographers. Because it requires a lot of skills and experience to interactively operate elementary photo enhancement tools (e.g., white balance, contrast, saturation and so on) step-by-step to achieve expected adjustments. Due to its complexity, it is also time-consuming to perform manual adjustments for a large number of photos.
\begin{figure}[htbp] \label{example-fig}
\centering
\includegraphics[width=3.2in]{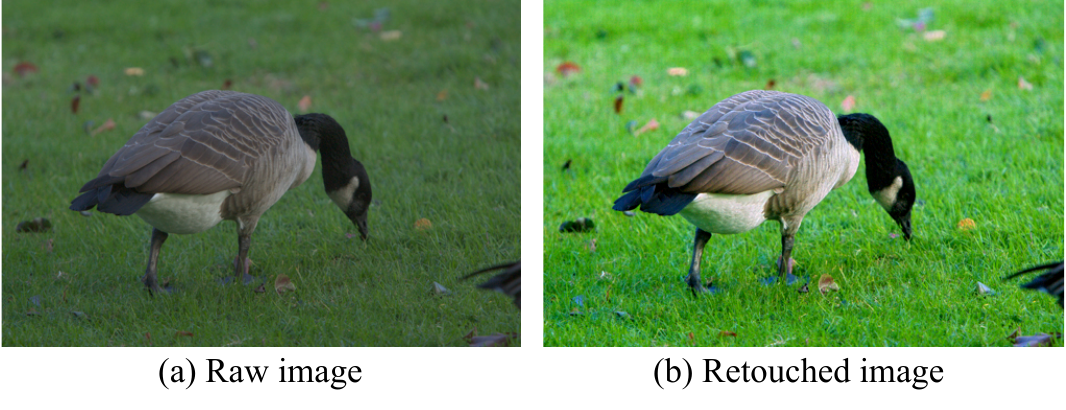}
\vspace{-1em}
\caption{\textbf{Example result.} (a) is a raw image. (b) is the corresponding retouched result of our model.}
\vspace{-1.3em}
\end{figure}

To address above problems, automatic color enhancement techniques are proposed to adaptively map photos to the satisfying styles and tones (as an example in figure 1). The challenge of the task is that the optimal mapping of a pixel is usually highly non-linear and dependents on not only the global color distribution of image but also local and contextual information \cite{hwang2012context,yan2016automatic,Chen:2018:DPE}. That is, the mapping from the raw image to the enhanced result is non-linear and spatially-varying.

There have been some works devoted to automatic color enhancement. \cite{fivek} built a dataset of 5,000 example input-output pairs and trained a global adjustment model on it. \cite{hwang2012context} proposed a local adjustment method that finding candidate images in dataset and searching for the best transformation of each pixel. \cite{learn-rank} proposed a learning-to-rank method to enhance images step-by-step like human photographers. 

More recently, deep learning techniques show powerful capacity in some vision tasks \cite{krizhevsky2012imagenet,he2016deep,wang2018non}, it also brings huge improvement to image color enhancement by learning from large amounts of paired raw-retouched images like the dataset MIT-Adobe-FiveK \cite{fivek}. Within these methods, \cite{yan2016automatic} used a fully-connected network to learn the transformation of each pixel with hand-craft feature. \cite{chen2017fast} proposed to learn image processing operators through fully convolutional network. 
\begin{figure*}[!htbp] \label{overview}
\centering
\includegraphics[width=7in]{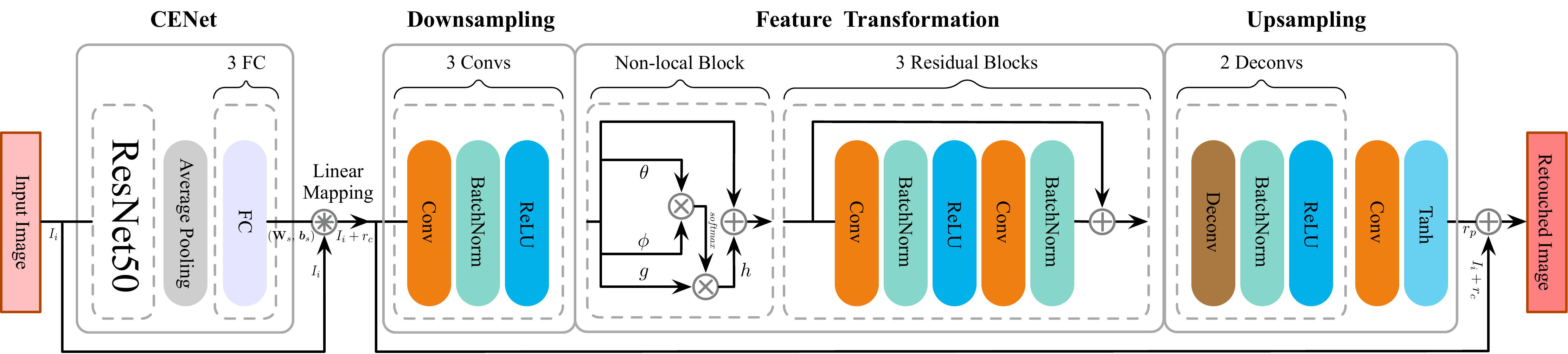}
\vspace{-1.2em}
\caption{\textbf{Overview of proposed coarse-to-fine framework.} The enhancement process is divided into channel-wise enhancement and pixel-wise refinement. In channel-wise enhancement, CENet predicts a 12-dim vector (i.e., ($\bm{\mathrm{W}}_s$, $\bm{b}_s$)) as parameters of linear mapping for RGB channels of input image. Then pixel-wise refinement is performed by predicting a refined residual image with PRNet. PRNet includes 3 stages: downsampling, feature transformation and upsampling.}
\vspace{-1em}
\end{figure*}

In addition, regarding color enhancement as image-to-image translation task \cite{pix2pix2017}, conditional Generative Adversarial Networks (GANs) \cite{goodfellow2014generative,mirza2014conditional} like \cite{pix2pix2017,CycleGAN2017} can also be applied to transforming raw images into retouched images. Based on this, \cite{ignatov2017dslr} collected aligned image pairs of the same scene by phone camera and DSLR camera, and trained a GAN model to learn the mapping between paired images. \cite{Chen:2018:DPE} proposed several improvements of current GAN models for image enhancement. \cite{deng2018aesthetic} proposed an aesthetic-driven image enhancement model by adversarial learning. These GAN-based methods are also effective for color enhancement. However, generating high-resolution retouched images with realistic effects is still a challenge \cite{Park_2018_CVPR,DeepExposure_NIPS}.

Moreover, Deep Reinforcement Learning (DRL) also makes contributions to image enhancement. For instances, \cite{DeepExposure_NIPS} proposed to learn local exposures with deep reinforcement adversarial learning. \cite{Park_2018_CVPR} proposed an easily interpretable step-by-step enhancement method. The authors treated the task as a Markov Decision Process (MDP), and used DRL algorithm to learn a sequence of global elementary enhancement actions. Global enhancement methods like \cite{Park_2018_CVPR} are effective to adjust images to states with global harmonious perception. However, in such global methods, the adjustments of every pixel is independent on pixel's position, preventing spatially-varying adjustment. Compared to them, pixel-wise enhancement methods have more flexible local adjustment but can also lead to local artifacts especially for handling high-solution images \cite{Park_2018_CVPR}, which breaks long-range perceptual color consistency and causes terrible global perceptual style. In view of this observation, in this paper, we propose a coarse-to-fine model with non-local attention for 
automatic color enhancement.

In our model, we propose to divide automatic color enhancement process into two mappings: a channel-wise enhancement and a pixel-wise refinement. We learn each mapping using a CNN. In this way, channel-wise enhancement performs global adjustment and generate coarse enhanced results with global harmonious perception without causing extra complex local artifacts. After that, pixel-wise refinement will be applied for local adjustment. Through such design, we learn refined residual images to adjust local details over our coarse enhanced results instead of directly learning the mapping from the original image to the target image. It therefore keeps global harmonious perception and better details in final results. In addition, another important innovation in this paper is to apply non-local attention blocks \cite{wang2018non} within our pixel-wise refinement network, which helps maintain long-range perceptual color consistency by capturing long-distance context information from global information for local adjustment. The ablation experiments indicate the effectiveness of two proposed enhancers, and the performances of proposed model outperform up-to-date works in quantitative or qualitative comparisons.

This paper is organized as follows: In Section 2, we describe our methods in detail. In Section 3, we present experiments and analysis. In Section 4, we conclude this work.
\vspace{-0.3em}
\section{The proposed method}
\label{sec:method}
\vspace{-0.5em}
Our coarse-to-fine model consists of two enhancers: a channel-wise enhancement network (named CENet) for global adjustment and a pixel-wise refinement network (named PRNet) for local adjustment. We adopt residual learning \cite{kim2016accurate} to train these two CNNs, which is efficient for tasks that input images and ground truth are largely similar \cite{Chen:2018:DPE,kim2016accurate}. Our model can be formulated as:
\vspace{-0.2em}
\begin{equation} \label{eq:model}
I_r = I_i + r_c + r_p ,  
\end{equation}
\vspace{-0.2em}
where $I_r$ and $I_i$ are retouched image and original input image, respectively. $r_c$ is the residual image predicted by 
CENet, and $r_p$ is another residual image predicted by PRNet.
\vspace{-1.8em}
\subsection{Channel-wise Enhancement}
\label{ssec:CE}
\vspace{-0.4em}
In \cite{Park_2018_CVPR}, a sequence of linear arithmetic is operated
on RGB channels step-by-step. Suppose $I_n$ is the output image after $n$-step linear mapping. We conclude their step-by-step enhancing process as:
\vspace{-0.2em}
\begin{equation}
\label{eq:step-by-tep}
\bm{p}_{n+1}=\bm{\mathrm{W}}_i\bm{p}_{n} + \bm{b}_i, \quad n=0, 1, 2, \cdots, N, \quad (\bm{\mathrm{W}}_i, \bm{b}_i) \in \mathcal{A},
\end{equation}
where $\bm{p}_{n}$ is a $3$-dim vector, it represents arbitrary pixel
 in image $I_n$. $\bm{\mathrm{W}}_i$ and $\bm{b}_i$ are a $3\times 3$ weight matrix and a $3$-dim bias vector, respectively. The tuple ($\bm{\mathrm{W}}_i$, $\bm{b}_i$) defines a unique linear enhancement operation as an element of discrete finite enhancement operation set $\mathcal{A}$. For example, to decrease the brightness with $0.05$, $\bm{\mathrm{W}}_i$ is set to a diagonal matrix given by diag$(0.95, 0.95, 0.95)$, and $\bm{b}_i$ is set to $\bm{0}$.

We consider that a sequence of linear enhancement operations actually is equivalent to a single direct linear mapping from the original image $I_0$ to the final result $I_N$. Thus, instead of learning the operation sequence by DRL \cite{Park_2018_CVPR}, we directly learn a single channel-wise linear mapping ($\bm{\mathrm{W}}_s$, $\bm{b}_s$) with a neural network (i.e., CENet) by end-to-end training. As presented in figure 2, we adopt CENet consists of ResNet50 \cite{he2016deep} (removed classifier layers) with 3 fully-connected layers to predict ($\bm{\mathrm{W}}_s$, $\bm{b}_s$) for each input image. Unlike finite and fixed enhancement operations in \cite{Park_2018_CVPR}, we predict the elements of $\bm{\mathrm{W}}_s$ and $\bm{b}_s$ in continuous space, which can be considered as an extending of $\mathcal{A}$ and provides more flexible and general global adjustments. After that $r_c$ can be calculated by ($\bm{\mathrm{W}}_s$, $\bm{b}_s$) and $I_i$ as follows:
\begin{equation}
\label{eq:s}
\bm{p}_{c}=\bm{\mathrm{W}}_s\bm{p}_{i} + \bm{b}_s
\end{equation}
where $\bm{p}_{i}$ represents arbitrary pixel in image $I_i$, and $\bm{p}_{c}$ represents the pixel at the same location in $r_c$. 
The final channel-wise enhancement result is equal to $I_i + r_c$. Due to linear mapping is  differentiable, efficient end-to-end training can be applied by using mini-batch gradient decent \cite{lecun1989backpropagation} to minimize the MSE loss, which is defined as: 
\begin{equation} \label{eq:loss}
\argmin_{\theta_c} \frac{1}{MN}\sum^{M}_{m=1}\sum^{N}_{n=1}(I_i^{mn} + r_c^{mn} - I_r^{mn})^2,
\end{equation}
where $M,N$ are the width and height of images respectively, superscript $m, n$ represent column and row index of pixel respectively, $\theta_c$ represents parameters of CENet. 

We noticed that the piecewise enhancer in \cite{deng2018aesthetic} might look similar to ours, but there are differences between \cite{deng2018aesthetic} and our method. \cite{deng2018aesthetic} predicts parameters set for 3 piecewise functions, each of which formulates adjustment of one channel in CIELab color space. Their adjustment of each channel is irrelevant to the other 2 channels. Unlike \cite{deng2018aesthetic}, we predict parameters for channel-wise linear mapping, and the adjustment of each channel is equal to a linear combination of all 3 channels, which can be viewed as adjusting one channel by extra information in other 2 channels.
\vspace{-0.5em}
\subsection{Pixel-wise Refinement}
\label{ssec:PR}
\vspace{-0.5em}
In pixel-wise refinement, we adopt PRNet to predict color residual $r_p$, which can be formulated as:
\vspace{-0.2em}
\begin{equation} \label{eq:pr}
r_p = \Phi(I_i + r_c),
\end{equation}
\vspace{-0.2em}
where $\Phi$ denotes the mapping of PRNet. Similar to equation \ref{eq:loss}, the PRNet is also trained by using mini-batch gradient decent to minimize the MSE between $r_p$ and $I_r-(I_i + r_c)$.
 
Our PRNet has similar basic architecture to networks in \cite{CycleGAN2017,johnson2016perceptual}. It  can be divided into 3 stages as presented in figure 2. First, the downsampling stage includes one stride-1 convolution and two stride-2 convolutions. Each convolution layer is followed by a batch normalization layer \cite{ioffe2015batch} and a ReLU \cite{nair2010rectified}. We set the number of feature maps in first convolution layer to 16. Second, the feature transformation stage consists of one non-local block \cite{wang2018non} and three residual blocks \cite{he2016deep}. The non-local operation in non-local blocks utilizes all elements in input features while convolution only sums up weighted features in local receptive fields. Therefore, the non-local block can combine both non-local and local information \cite{wang2018non}, which helps PRNet capture more long-range dependencies from global features for pixel-wise adjustment and improve the performance. In practice, we attempted to place the non-local block in different locations, experiment results indicated placing it at the front of the stage could lead to relatively better performances. Third, the upsampling stage is symmetric with downsampling stage, it includes two stride-2 deconvolutions and one stride-1 convolution which can restore features to RGB channels. 

\begin{figure*}[!htbp] \label{results-fig}
\centering
\includegraphics[width=7in]{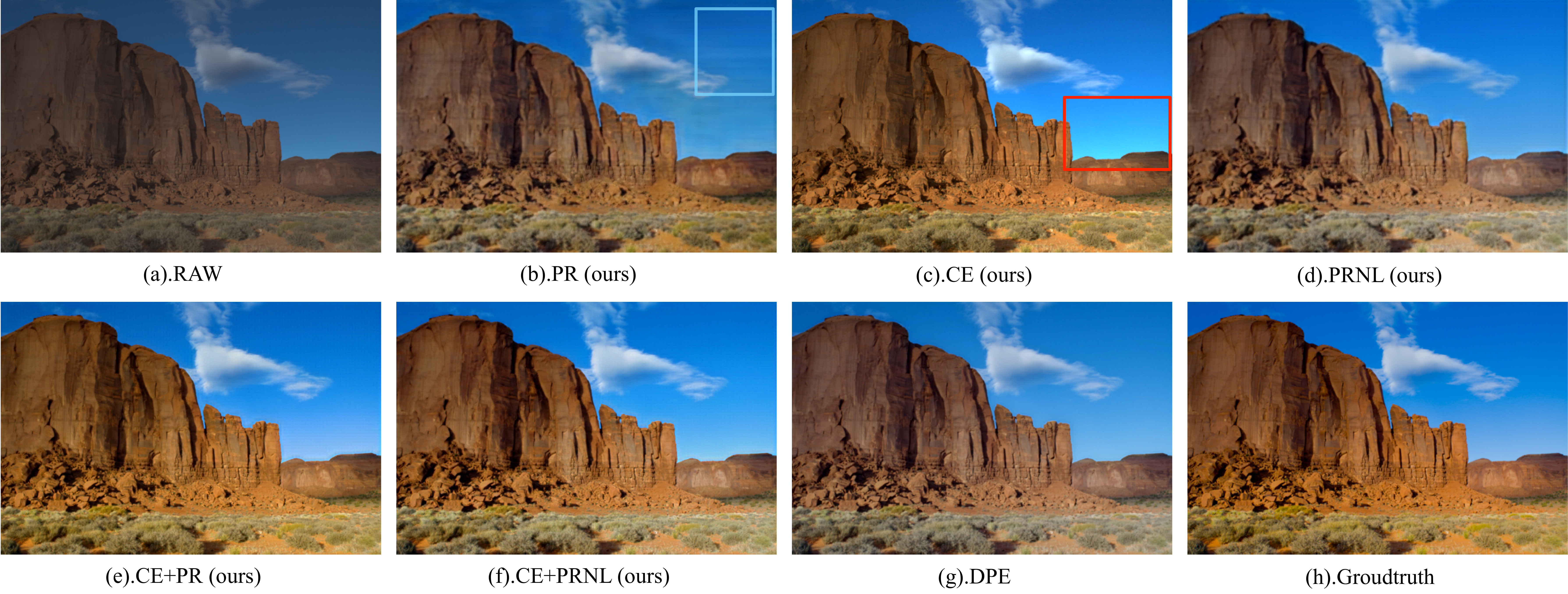}
\vspace{-2em}
\caption{\textbf{Qualitative comparisons of different results.} (a) and (h) are raw image and ground truth, respectively. (b)-(f) are the results of our models. (g) is the result of DPE\cite{Chen:2018:DPE}.}
\vspace{-0.8em}
\end{figure*}
\vspace{-0.5em}
\section{Evaluation}
\label{eva}
\vspace{-0.5em}
In this section, we first elaborate some implementation details and experiment settings for ablation study. We then describe our experimental results and the corresponding analysises.
\vspace{-0.6em}
\subsection{Dataset}
\label{ssec:DT}
\vspace{-0.5em}
We evaluate our method on dataset MIT-Adobe FiveK \cite{fivek}, which consists of 5000 raw images and each raw image was enhanced by 5 professional photographers. Following the common practice, we select the results of photographer C as the label and validate performances on the RANDOM250 \cite{hwang2012context,yan2016automatic,Park_2018_CVPR} which is a subset of MIT-Adobe FiveK. The training set consists of the rest 4750 pairs images. We keep the width-to-height ratio of images and resize them to 500 pixels on the longer edge.  
\vspace{-0.6em}
\subsection{Implementation Details}
\label{ssec:Imple-Detail}
\vspace{-0.5em}
For training CENet and PRNet, image pairs are padded to $500 \times 500$,  so the network can adjust arbitrary images with edges no longer than 500 pixels. Image values are normalized to [0, 1]. We adopt SGD optimizer with momentum of 0.9. The learning rate is initialized to 0.01 and reduced by a factor of 0.1 at every 10k steps. We set the batch size to 16 and stop the training after 200 epochs.
\vspace{-0.6em}
\subsection{Ablation Experiments}
\label{ssec:AE}
\vspace{-0.5em}
To evaluate every module in our coarse-to-fine framework, we conduct a series of ablation experiments as below:
\vspace{-0.5em}
\begin{itemize}[leftmargin=1em]
\setlength{\itemsep}{2pt}
\setlength{\parsep}{1em}
\setlength{\parskip}{0em}
\setlength{\itemindent}{0 em}
\item CE: This method adjusts raw images by channel-wise linear mapping using our CENet.
\item PR: Directly enhance the raw images using PRNet with 18 residual blocks in feature transformation stage.
\item PRNL: The same as PR but with 1 non-local block and 3 residual blocks in feature transformation stage.
\item CE+PR: Coarse-to-fine method with both CENet and PRNet with 3 residual blocks in feature transformation stage.
\item CE+PRNL:The same as CE+PR except an additional non-local block in feature transformation stage of PRNet.
\end{itemize}
\vspace{-1em}
\subsection{Results Analysis}
\label{results} 
\vspace{-0.3em}
\noindent \textbf{Ablation Study.} We first evaluate all the methods defined in section 3.3. Table 1 shows quantitative results of those methods. Like previous methods  \cite{hwang2012context,yan2016automatic,Chen:2018:DPE,Park_2018_CVPR}, we compute $L^2$ error in CIELab color space and PSNR to compare numerical magnitude of enhancements. SSIM is measured to quantitatively compare local artifacts \cite{Chen:2018:DPE,Park_2018_CVPR}. From the results, we can conclude as follows:
\begin{table}[!ht]  \label{tab:table1}
 \centering
\begin{tabular}{c|c|c|c}
\hline
method & $L^2$ error (LAB) & PSNR & SSIM \\
\hline
\hline
CE & 10.32 & 22.85 & 0.893 \\
PR & 10.93 & 22.07 & 0.882 \\
PRNL & 9.32 & 23.90 & 0.905 \\
CE+PR & 9.50 & 23.89 &  0.906\\
CE+PRNL & \textbf{9.10} & \textbf{24.19} & \textbf{0.915} \\
\hline
\end{tabular}
\vspace{-0.5em}
\caption{Quantitative performances of our methods on RANDOM250 \cite{hwang2012context,yan2016automatic,Park_2018_CVPR}.}
\end{table}
\vspace{-1em}
\begin{enumerate}[label=\arabic*., leftmargin=1em]
\setlength{\itemsep}{3pt}
\setlength{\parsep}{1em}
\setlength{\parskip}{0em}
\setlength{\itemindent}{0 em}
\item CE+PR outperforms both CE and PR a lot on all quantitative metrics. This is because that CE lacks local adjustments (see red box in figure 3(c)) although its results have close global color style to the ground truth. And PR is likely to cause complex artifacts due to its flexible local adjustment, its results are more blurred with local artifacts (see blue box in figure 3(b)). However, our coarse-to-fine framework CE+PR can keep harmonious global color style and refined details in final results, which leads to quantitative and qualitative improvements.
\item PRNL also outperforms PR. Its performances are close to CE+PR. This indicates that the non-local attention block in PRNet can provide a different way to reduce artifacts by capturing long-range dependencies between features. 
\item CE+PRNL achieves best performance of our methods. It improves the $L^2$ error in CIELab color space by around 0.18, improves the PSNR by around 0.3 dB, and improves the SSIM by around 0.01 compared with CE+PR and PRNL, which indicates that both embedding non-local block in CE+PR or adding CE in front of PRNL can offer further improvements.
\end{enumerate}

\begin{table}[ht]  \label{tab:table2}
\vspace{0.6em}
\centering
\begin{tabular}{c|c|c|c}
\hline
method & $L^2$ error (LAB) & PSNR & SSIM \\
\hline
\hline
Exemplar-based \cite{hwang2012context} & 15.01 & - & - \\
DeepNet-based \cite{yan2016automatic} & 9.85 & - & - \\
DeepRL-based \cite{Park_2018_CVPR} & 10.99 & - & 0.905 \\
CE (ours) & 10.32 & 22.85 & 0.893 \\
DPE \cite{Chen:2018:DPE} & 9.93 & 23.89 &  0.906\\
CE+PRNL (ours) & \textbf{9.10} & \textbf{24.19} & \textbf{0.915} \\
\hline
\end{tabular}
\vspace{-0.5em}
\caption{Quantitative performances comparisons of different methods on RANDOM250 \cite{hwang2012context,yan2016automatic,Park_2018_CVPR}.}
\vspace{-0.5em}
\end{table}
\noindent \textbf{Benchmark Comparison.} 
In table 2, we compare quantitative performances with other methods. We tested DPE\cite{Chen:2018:DPE} on the same dataset with official implementation. In comparisons of channel-wise enhancements, our CE delivers better performance on $L^2$ error than DeepRL-based due to our channel-wise linear enhancement operations are more flexible compared with fixed operations in DeepRL-based \cite{Park_2018_CVPR}. As for pixel-wise retouching results, our CE+PRNL has better quantitative performances than \cite{hwang2012context,yan2016automatic,Park_2018_CVPR,Chen:2018:DPE}. In figure 3, we can see that the results of ours are also reasonable compared with the ground truth and competitive to DPE \cite{Chen:2018:DPE}. All these results indicate our model is effective to image color enhancement.
\vspace{-1em}
\section{Conclusion}
\label{sec:conclu}
\vspace{-0.5em}
In this work, we propose a coarse-to-fine automatic color enhancement framework, which consists of channel-wise enhancement and pixel-wise refinement. In channel-wise enhancement, we learn a linear mapping for RGB channels. In pixel-wise refinement, refined residual images are predicted for local adjustments. Experimental results demonstrate that each component in our framework is effective to improve the final performance. In addition, our fully-equipped model outperforms related methods on the benchmark.
\bibliographystyle{IEEEbib}
\bibliography{strings,refs}

\end{document}